\documentclass[11pt]{article}
\usepackage[table]{xcolor}

\usepackage[final]{acl}

\usepackage{times}
\usepackage{latexsym}
\usepackage{amsmath}
\usepackage{amssymb}
\usepackage{pifont}
\usepackage{algorithm}
\usepackage{algpseudocode}
\usepackage{booktabs}
\usepackage{multirow}

\usepackage[T1]{fontenc}

\usepackage[utf8]{inputenc}

\usepackage{microtype}

\usepackage{inconsolata}

\usepackage{graphicx}

\definecolor{myred}{HTML}{E74C3C}
\definecolor{myblue}{HTML}{3498DB}
\definecolor{mygreen}{HTML}{27AE60}
\definecolor{myorange}{HTML}{ff8800}
\definecolor{mypurple}{HTML}{9B59B6}
\definecolor{mypink}{HTML}{f7acb9}

\definecolor{headercolor}{RGB}{100,100,100}
\definecolor{bluecolor}{RGB}{0,100,200}
\definecolor{orangecolor}{RGB}{200,100,50}
\definecolor{redcolor}{RGB}{200,0,0}

\newcommand{\red}[1]{\mytextbox{myred}{\textbf{\textcolor{myred}{#1}}}}
\newcommand{\blue}[1]{\mytextbox{myblue}{\textbf{\textcolor{myblue}{#1}}}}
\newcommand{\green}[1]{\mytextbox{mygreen}{\textbf{\textcolor{mygreen}{#1}}}}

\newcommand{\purple}[1]{\mytextbox{mypurple}{\textbf{\textcolor{mypurple}{#1}}}}

\usepackage{tcolorbox}
\newcommand{\mytextbox}[2]{%
  \colorbox{#1!15}{\strut #2}%
}

\title{Efficient Agentic Reinforcement Learning with On-Policy Intrinsic Knowledge Boundary Enhancement}

\author{Dingwei Chen{$^{\spadesuit\Diamond}$\thanks{~~Work was done during the internship at Tencent Inc.}}, Zefang Zong{$^{\spadesuit}$}, Zhipeng Ma{$^{\spadesuit}$}, Leo Luo{$^{\spadesuit}$}, Yang Li{$^{\spadesuit}$} \\
  \textbf{Chengming Li}{$^{\heartsuit}\thanks{~~Corresponding author.}$}, \textbf{Peng Chen}{$^{\spadesuit}$}, \textbf{Jie Jiang}{$^{\spadesuit}\footnotemark[2]$} \\
  $^\spadesuit$Tencent Inc   $^\Diamond$The Chinese University of Hong Kong  $^\heartsuit$Shenzhen MSU-BIT University\\
  \texttt{cuso4cdw@gmail.com}, \texttt{licm@smbu.edu.cn} \\
  \texttt{\{willzong,thomasyngli\}@tencent.com}\\
  }

\begin{document}
\maketitle
\begin{abstract}
Agentic reinforcement learning (RL) has proven effective for training LLM-based agents with external tool-use capabilities. However, we identify that agentic RL training induces increasing redundant tool calls and blurs the model's intrinsic knowledge boundary, where the model fails to distinguish when tools are needed versus when parametric knowledge suffices. Existing solutions based on reward shaping create coarse-grained optimization targets that tend to incentivize indiscriminate tool-call suppression, leading to reward hacking. In this paper, we propose \textbf{AKBE} (\textbf{\underline{A}}gentic \textbf{\underline{K}}nowledge \textbf{\underline{B}}oundary \textbf{\underline{E}}nhancement), an on-policy method that dynamically probes the model's intrinsic knowledge boundary through dual-path (with-tool and no-tool) rollouts during training. We define the knowledge boundary as the per-instance determination of whether tools are required and the minimum tool calls necessary. By comparing correctness across paths, AKBE categorizes trajectories and constructs targeted supervisory signals that guide efficient tool-use patterns for each question. These signals are integrated seamlessly into the agentic RL training loop. Experiments on seven QA benchmarks demonstrate that AKBE improves task accuracy by $+1.85$ on average and reduces tool calls by 18\% over standard agentic RL, yielding 25\% higher tool productivity without any accuracy-efficiency trade-off. Further analysis suggests its plug-and-play compatibility across different RL algorithms and the mechanism of each signal category. Our code is available at \href{https://github.com/CuSO4-Chen/AKBE}{\texttt{https://github.com/CuSO4-Chen/AKBE}}.
\end{abstract}

\begin{figure}[t]
  \centering
  \includegraphics[width=1.0\linewidth]{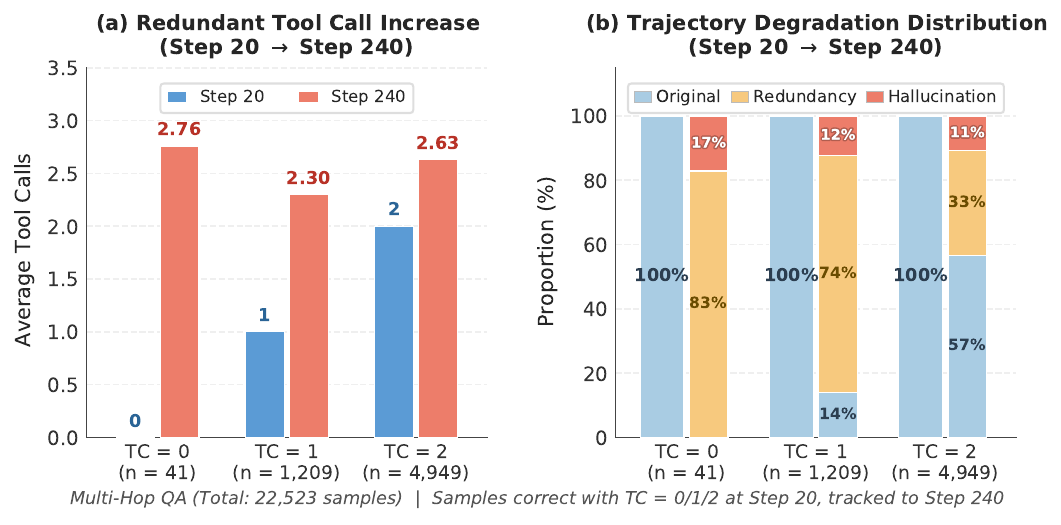}
  \caption{Redundant tool-call growth during GRPO training (Qwen3-4B Multi-Hop). Samples correctly answered at early training (Step 20) with TC = 0/1/2 are tracked to late training (Step 240). \textit{Left}: Tool calls increase substantially across all groups. \textit{Right}: Trajectory degradation into \textbf{original} (still correct), \textbf{redundant} (correct but with extra TC), and \textbf{hallucinated} (degraded to incorrect due to noisy retrieval) categories.}
  \label{fig:intro}
\end{figure}

\section{Introduction}
Large language model (LLM) agents have demonstrated remarkable capabilities in solving complex tasks by integrating internal reasoning with external tool interactions \citep{yao2023react, schick2023toolformer, si2026context, luo2026tabtracer}. Using tools such as search engines and code interpreters, these agents extend their reasoning beyond parametric knowledge. Recently, reinforcement learning has emerged as a powerful post-training paradigm for further enhancing agentic capabilities, with methods such as GRPO \citep{shao2024deepseekmathpushinglimitsmathematical}, DAPO \citep{yu2025dapo}, and specialized agentic RL algorithms \citep{feng2025group, dong2025agenticaepo, zong2026at2po} achieving promising improvements on tool-augmented reasoning benchmarks.


However, a critical yet underexplored side effect of agentic RL training is that: as the model is optimized to enhance reasoning capability with tool access, it increasingly produces redundant tool calls, either invoking tools when parametric knowledge suffices or making excessive calls when fewer would suffice, which is defined as \textbf{\textit{cognitive offloading}} \citep{wang2025acting, xie2026over}. This manifests itself as a steady growth in tool calls during training, as illustrated in Figure~\ref{fig:intro}. Such an over-reliance on tool calls is problematic in two ways: (1) it wastes computational resources and increases inference latency; and (2) unnecessary tool calls may introduce noise that overrides correct internal reasoning with misleading retrieved information, leading to degradation of answer quality.

Existing approaches to efficient agentic RL address this issue primarily through reward shaping, incorporating tool-call patterns into the reward function \citep{wang2025acting, wu2025search}. However, directly coupling tool-call behavior with reward signals creates a coarse-grained optimization target. This incentivizes the model to reduce overall tool usage to gain extra reward regardless of whether specific calls are necessary, leading to reward hacking and degraded task accuracy. More fundamentally, such reward-level approaches cannot capture the per-instance distinction between necessary and redundant tool calls, nor adapt to the dynamic evolution of the model's knowledge boundary throughout training.

In this paper, we propose \textbf{AKBE} (\textbf{\underline{A}}gentic \textbf{\underline{K}}nowledge \textbf{\underline{B}}oundary \textbf{\underline{E}}nhancement), an on-policy method that addresses this limitation by explicitly probing the model's intrinsic knowledge boundary during training. We define the \textbf{\textit{knowledge boundary}} as the per-instance determination of whether external tools are required and, when required, the minimum tool invocations necessary to reach the correct answer, representing the most efficient tool-call pattern for each question. The key insight is that for each question in a training batch, we perform dual-path rollouts with and without external tools. By comparing the correctness of these two paths, we identify whether a question lies within the model's parametric knowledge or genuinely requires external tool calls, and further determine the minimum tool usage required in the latter case. Based on this identification, AKBE categorizes each question and constructs targeted supervisory signals: \textbf{\textit{Tool-dependent}} selects minimum tool-call correct trajectories to reinforce efficient tool use, \textbf{\textit{Efficiency}} selects no-tool correct trajectories to eliminate redundant calls, \textbf{\textit{Hallucination}} selects no-tool correct trajectories to alleviate harmful tool reliance, and \textbf{\textit{Both-wrong}} provides no signal, relying solely on the RL objective. These knowledge boundary-guided signals are integrated seamlessly into the training loop with the standard RL objective as an auxiliary on-policy training loss, providing fine-grained instance-level guidance without modifying the RL reward or optimization process. Our contributions are summarized as follows:

\begin{itemize}
    \item We propose AKBE, an on-policy knowledge boundary enhancement method for efficient agentic RL that dynamically probes the model's intrinsic knowledge boundary through dual-path rollouts and constructs boundary-guided supervisory signals to eliminate redundant tool calls and reinforce efficient tool-use patterns.
    \item We conduct extensive experiments on seven QA benchmarks across two backbone models, demonstrating that AKBE improves task accuracy by $+1.85$ on average and reduces tool calls by 18\% over standard agentic RL, yielding 25\% higher tool productivity. It outperforms baseline methods in most cases without any accuracy-efficiency trade-off.
    \item We further demonstrate that AKBE serves as a plug-and-play module compatible across diverse agentic RL algorithms, and reveal that the model's knowledge boundary evolves dynamically during training, where each signal category naturally adapts to address a distinct failure mode of tool-use behavior.
\end{itemize}

\section{Related Work}
\label{sec:related_work}

Recent work applies reinforcement learning to train LLM-based agents with external tool-use capabilities \citep{shao2024deepseekmathpushinglimitsmathematical, yu2025dapo, zheng2025groupsequencepolicyoptimization}. Furthermore, a series of work designs specialized algorithms tailored to agentic settings such as entropy-driven rollout and credit assignment \citep{jin2025search, dong2025agenticaepo, ji2025treesearchllmagent, zong2026at2po, chen2026a2tgpo}. However, these methods all exhibit increasing redundant tool calls during training \citep{xie2026over}. To mitigate this, OTC-PO \citep{wang2025acting} introduces a tool-productivity reward term, $\beta$-GRPO \citep{wu2025search} incorporates confidence thresholds, and HiPRAG \citep{wu2025hiprag} applies hierarchical process rewards to evaluate the tool-call of each step. However, these reward-based methods either apply coarse-grained penalties on overall tool-call behavior where agents always learn to reduce tool calls indiscriminately to gain extra reward, leading to reward hacking, or evaluate each tool-call step individually but rely on external models or APIs \citep{wu2025hiprag}, introducing additional overhead and dependencies. SMART \citep{qian2025smart} instead constructs metacognitive SFT data offline, but static datasets cannot track the evolving knowledge boundary during RL training. Unlike these approaches, our proposed AKBE operates within the RL training loop, dynamically probing the model's intrinsic knowledge boundary via on-policy dual-path (with-tool and no-tool) rollouts to construct boundary-guided supervisory signals that seamlessly integrate with any agentic RL algorithm as a plug-and-play module.

\section{Preliminary}
\label{sec:preliminary}

\subsection{Task Definition}

We consider an agentic setting where a language model policy $\pi_\theta$ iteratively interacts with an external tool environment $E$ to answer a given question $q$. Following the ReAct paradigm \citep{yao2023react}, the agent generates a sequence of interleaved reasoning-and-action turns. At each turn $t$, the agent produces a thought and an action $a_t$ conditioned on the current context $c_t$. The action is either an invocation of an external tool, which returns an observation $o_t$ appended to the context, or a \texttt{finish} action that terminates the episode and returns the final answer. A complete interaction thus forms a trajectory $y = (a_1, o_1, \ldots, a_T)$, where $T$ denotes the final step. An outcome reward $R(y)$ is assigned based on whether the final answer matches the ground truth. The learning objective is to maximize the expected reward over the training distribution $\mathcal{D}$:
\begin{equation}
    J(\pi_\theta) = \mathbb{E}_{q \sim \mathcal{D}} 
    \mathbb{E}_{y \sim \pi_\theta(\cdot | q, E)}   \left[ 
    R(y) \right]
\end{equation}

\subsection{Agentic Reinforcement Learning}

While PPO \citep{schulman2017proximal} provides a general policy optimization framework, its reliance on a separate value evaluator introduces substantial memory and training overhead. GRPO \citep{shao2024deepseekmathpushinglimitsmathematical} addresses this by introducing the group-relative advantages, and has become the predominant algorithm in recent agentic RL research \citep{jin2025search, dong2025agenticaepo, ji2025treesearchllmagent}.

Specifically, for each question $q$, GRPO samples a group of $G$ trajectories $\{y_i\}_{i=1}^G$ from the current policy $\pi_\theta$ and computes group-relative advantages:
\begin{equation}
    \hat{A}_i = \frac{R(y_i) - \text{mean}(\{R(y_j)\}_{j=1}^G)}{\text{std}(\{R(y_j)\}_{j=1}^G)}
\end{equation}
The policy is updated by maximizing the clipped policy objective with a KL regularization term:
\begin{align}
    \mathcal{L}_{\text{GRPO}} =& -\frac{1}{G}\sum_{i=1}^{G} \Big[ \min\big( r_i(\theta) \hat{A}_i, \; \text{clip}(r_i(\theta), \notag \\ & 1\!-\!\epsilon, 1\!+\!\epsilon) \hat{A}_i \big)  - \beta \, D_{\text{KL}}(\pi_\theta \| \pi_{\text{ref}}) \Big]
\end{align}
where $r_i(\theta) = \frac{\pi_\theta(y_i | q)}{\pi_{\theta_{\text{old}}}(y_i | q)}$ is the importance sampling ratio, $\epsilon$ is the clipping threshold, and $\beta$ controls the strength of KL regularization against a reference policy $\pi_{\text{ref}}$. Note that tokens from tool observation are masked out during training.


\begin{figure*}[t]
  \centering
  \includegraphics[width=1.0\linewidth]{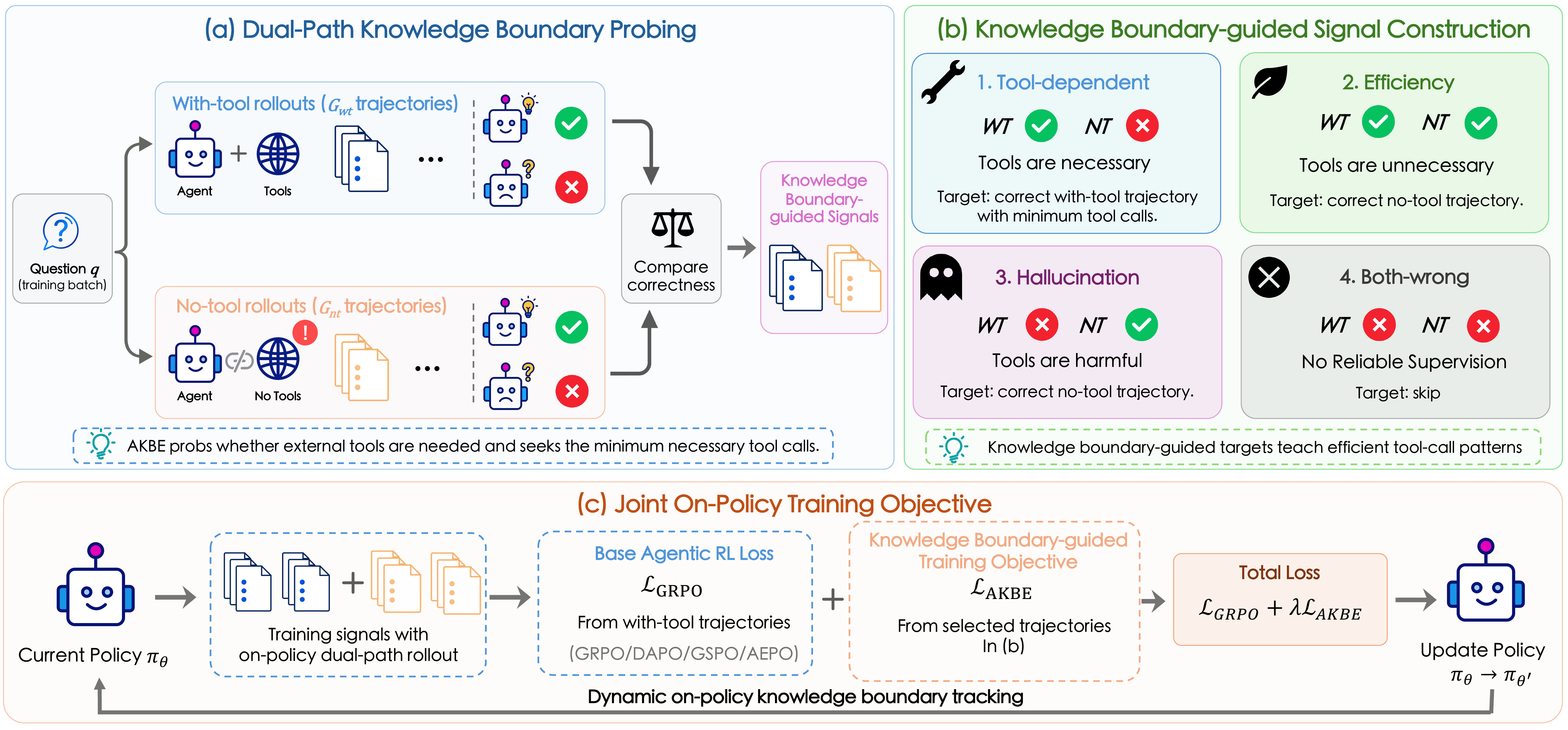}
  \caption{The framework of AKBE. For each question, dual-path rollouts (with-tool and no-tool) are performed in parallel. Based on the correctness of each path, corresponding target trajectories are   
  selected to construct on-policy knowledge boundary-guided supervisory signals. These signals are integrated with the agentic RL objective.}
  \label{fig:method}
\end{figure*}

\begin{algorithm}[t]
\caption{AKBE Training (per batch)}
\label{alg:akbe}
\begin{algorithmic}[1]
\Require Training batch $\mathcal{B}$, policy $\pi_\theta$, with-tool rollout count $G_{wt}$, no-tool rollout count $G_{nt}$, coefficient $\lambda$
\State $\mathcal{S} \leftarrow \emptyset$
\For{each question $q \in \mathcal{B}$}
    \State Sample $G_{wt}$ with-tool trajectories $\{y_q^{(i)}\}_{i=1}^{G_{wt}}$ from $\pi_\theta$
    \State Sample $G_{nt}$ no-tool trajectories $\{\hat{y}_q^{(i)}\}_{i=1}^{G_{nt}}$ from $\pi_\theta$
    \State $WT \leftarrow 1$ if $\exists \, y_q^{(i)}$ s.t. $R(y_q^{(i)}) = 1$, else $0$
    \State $NT \leftarrow 1$ if $\exists \, \hat{y}_q^{(i)}$ s.t. $R(\hat{y}_q^{(i)}) = 1$, else $0$
    \If{$WT \wedge \neg NT$} \Comment{Tool-dependent}
        \State $y_q^* \leftarrow \arg\min_{\text{correct } y_q^{(i)}} \text{TC}(y_q^{(i)})$
        \State $\mathcal{S} \leftarrow \mathcal{S} \cup \{(q, y_q^*)\}$
    \ElsIf{$WT \wedge NT$} \Comment{Efficiency}
        \State $y_q^* \leftarrow \text{RandomSelect}(\text{correct } \hat{y}_q^{(i)})$
        \State $\mathcal{S} \leftarrow \mathcal{S} \cup \{(q, y_q^*)\}$
    \ElsIf{$\neg WT \wedge NT$} \Comment{Hallucination}
        \State $y_q^* \leftarrow \text{RandomSelect}(\text{correct } \hat{y}_q^{(i)})$
        \State $\mathcal{S} \leftarrow \mathcal{S} \cup \{(q, y_q^*)\}$
    \EndIf \Comment{Both-wrong: skip}
\EndFor
\State Compute $\mathcal{L}_{\text{GRPO}}$ from with-tool trajectories
\State $\mathcal{L}_{\text{AKBE}} \leftarrow - \sum_{(q, y_q^*) \in \mathcal{S}} \log \pi_\theta(y_q^* \mid q)$
\State Update $\pi_\theta$ with $\mathcal{L}_{\text{total}} = \mathcal{L}_{\text{GRPO}} + \lambda \cdot \mathcal{L}_{\text{AKBE}}$
\end{algorithmic}
\end{algorithm}

\section{Method}

In this section, we present AKBE, which augments the agentic RL objective with knowledge boundary-guided training signals derived from dual-path rollouts. By probing whether the model needs external tools for each question and how many calls are minimally required, AKBE selects efficient
trajectories as targeted on-policy optimization signals that eliminate redundant tool calls while reinforcing efficient tool use where external tools are genuinely needed. We illustrate the framework in Figure~\ref{fig:method} and detail the training procedure in Algorithm~\ref{alg:akbe}.

\subsection{Dual-Path Trajectory Rollout}

For each question $q$ in a training batch, AKBE performs a dual-path trajectory rollout (with-tool and no-tool) in parallel:

\textbf{With-tool trajectory rollout:} We sample $G_{wt}$ agentic rollouts where policy $\pi_\theta$ has access to external tools. Their trajectories consist of one or more tool calls. Let $WT$ denote whether at least one with-tool trajectory yields a correct answer.

\textbf{No-tool trajectory rollout:} We sample $G_{nt}$ rollouts in which tool access is disabled, forcing $\pi_\theta$ to rely solely on its parametric knowledge. Let $NT$ denote whether at least one no-tool trajectory yields a correct answer.

We define the \textit{knowledge boundary} of $\pi_\theta$ on question $q$ as:
\begin{align}
    \text{KB}(q, \pi_\theta) = \mathbb{1}\big[ \exists \, \hat{y}_q^{(i)} &\in \{\hat{y}_q^{(1)}, \ldots, \hat{y}_q^{(G_{nt})}\} \notag \\
    &\text{s.t. } R(\hat{y}_q^{(i)}) = 1 \big]
\end{align}
where $\text{KB}=1$ indicates that $q$ lies within the model's intrinsic knowledge (i.e., tool calls are unnecessary), and $\text{KB}=0$ indicates that external tools are required. Since the no-tool rollouts do not involve any tool interaction or environment latency, they incur substantially lower time consumption compared to with-tool rollouts, making this probing step computationally efficient.

\subsection{Boundary-Guided Signal Construction}
\label{sec:signal_construction}

Based on the dual-path outcomes $(WT, NT)$, we classify trajectories for each question into four categories and construct corresponding training signals:

\textbf{Tool-dependent ($WT$=\checkmark, $NT$=\ding{55}).} The model can only answer correctly with tool calls ($\text{KB}=0$), where tool calls are necessary. We select the correct with-tool trajectory with the \textit{minimum number of tool calls} as the target $y_q^*$, reinforcing efficient tool-use patterns while preserving necessary tool invocations. When multiple correct trajectories share the same minimum tool-call count, we randomly sample one to avoid bias. At a finer granularity, each tool invocation reflects a dynamic step-level knowledge boundary decision: the model invokes a tool when its parametric knowledge is insufficient for a specific process reasoning step. Selecting the minimum tool-call trajectory thus reinforces the broadest achievable knowledge boundary at each step for a specific question.

\textbf{Efficiency ($WT$=\checkmark, $NT$=\checkmark).} The model can answer correctly without tools ($\text{KB}=1$), making tool calls redundant. We randomly select a correct no-tool trajectory as the target $y_q^*$, teaching the model to bypass unnecessary tool invocations for questions within its knowledge boundary.

\textbf{Hallucination ($WT$=\ding{55}, $NT$=\checkmark).} The model answers correctly without tools but incorrectly with tools ($\text{KB}=1$), indicating that tool calls introduce harmful noise or lead the model towards erroneous reasoning paths. We select a correct no-tool trajectory as the target $y_q^*$, steering the model away from detrimental tool reliance for a specific question.

\textbf{Both-wrong ($WT$=\ding{55}, $NT$=\ding{55}).} Neither path yields a correct answer. No reliable supervisory signal can be constructed; we rely solely on the original RL objective for these instances.

\subsection{Joint Training Objective}

The overall training objective combines the original RL loss with the knowledge boundary-guided training objective:
\begin{equation}
    \mathcal{L}_{\text{total}} = \mathcal{L}_{\text{GRPO}} + \lambda \cdot \mathcal{L}_{\text{AKBE}}
\end{equation}
where $\mathcal{L}_{\text{GRPO}}$ can be replaced by any classic agentic RL loss (e.g., DAPO, GSPO), and $\mathcal{L}_{\text{AKBE}}$ is the on-policy cross-entropy training objective over the selected target trajectories:
\begin{equation}
    \mathcal{L}_{\text{AKBE}} = - \sum_{q \in \mathcal{S}} \log \pi_\theta(y_q^* \mid q)
\end{equation}
where $\mathcal{S} = \mathcal{S}_{\text{dep}} \cup \mathcal{S}_{\text{eff}} \cup \mathcal{S}_{\text{hal}}$ denotes the set of questions with constructed signals from the Tool-dependent, Efficiency, and Hallucination categories respectively, and $y_q^*$ is the selected target trajectory for question $q$ as described in \S\ref{sec:signal_construction}. The coefficient $\lambda$ controls the strength of the boundary-guided objective relative to the RL loss.

Crucially, since both $\mathcal{L}_{\text{GRPO}}$ and $\mathcal{L}_{\text{AKBE}}$ are computed from on-policy rollouts of the current $\pi_\theta$, the knowledge boundary is dynamically re-evaluated at every training step. As the model improves through RL training, the knowledge boundary for a specific question may shift, and the boundary-guided signal adapts accordingly. This on-policy nature distinguishes AKBE from approaches with static offline data which cannot track such dynamic evolution. Furthermore, AKBE is designed as a \textbf{plug-and-play} module: it can be seamlessly integrated with any agentic RL algorithm by simply adding the $\lambda \cdot \mathcal{L}_{\text{AKBE}}$ term during training regardless of the specific form of $\mathcal{L}_{\text{RL}}$.

\section{Experiments}

\begin{table*}[t]
  \centering                  
  \small   
  \setlength{\tabcolsep}{3.5pt}
  \renewcommand{\arraystretch}{1.1}
  \caption{Experiment results on two backbone models across seven datasets. The \textbf{bolded} values indicate the best result. Our proposed AKBE outperforms existing methods in most cases.}
  \label{tab:main_results}
  \begin{tabular}{l|ccccccc|cccccc}
  \toprule
   \multirow{2}{*}{\textbf{Method}} & \multicolumn{7}{c|}{\textbf{Multi-Hop}} & \multicolumn{6}{c}{\textbf{Single-Hop}} \\
  \cmidrule(lr){2-8}\cmidrule(lr){9-14}
   & Hotpot & 2Wiki & MuSiQue & Bamb. & Avg. & TC$\downarrow$ & TP$\uparrow$
                 & NQ & TriviaQA & PopQA & Avg. & TC$\downarrow$ & TP$\uparrow$ \\
  \midrule
  \multicolumn{14}{c}{\textit{Backbone Model: Qwen3-4B}} \\
  \midrule
  ReAct        & 30.42 & 32.92 & 12.83 & 44.80 & 30.01 & 3.43 & 8.75 & 26.75 & 53.53 & 35.34 & 41.31 & 2.27 & 18.19 \\
  Search-o1       & 35.18 & 38.45 & 14.21 & 48.00 & 34.83 & 3.31 & 10.52 & 33.85 & 57.42 & 40.27 & 46.11 & 1.72 & 26.81 \\
  R1-Searcher       & 45.62 & 49.83 & 19.51 & 50.40 & 45.20 & 3.28 & 13.78 & 46.85 & 64.52 & 49.43 & 54.95 & 1.48 & 37.13 \\
  Search-R1         & 44.80 & 50.66 & 18.94 & 52.00 & 45.40 & 3.16 & 14.33 & 45.98 & 65.17 & 49.18 & 54.97 & 1.41 & 38.99 \\
  OTC-PO       & 38.74 & 47.71 & 15.16 & 48.00 & 41.27 & \underline{2.06} & \underline{20.03} & 45.31 & 64.35 & 49.86 & 54.90 & \underline{1.02} & \underline{53.83} \\
  $\beta$-GRPO & 45.83 & 51.49 & 21.06 & 56.00 & 46.39 & 3.01 & 15.41 & 47.00 & 64.73 & 49.95 & 55.30 & 1.53 & 36.15 \\
  Offline AKBE & 45.21 & 51.12 & 19.84 & 54.40 & 45.84 & 2.45 & \textbf{18.71} & 46.52 & 64.55 & 49.72 & 55.06 & 1.22 & 45.13 \\
  \textbf{AKBE (Ours)} & \textbf{46.53} & \textbf{51.89} & \textbf{21.17} & \textbf{56.80} & \textbf{46.82} & 2.60 & 18.01 & \textbf{47.33} & \textbf{65.18} & \textbf{50.80} & \textbf{55.90} & 1.16 & \textbf{48.19} \\
  \midrule
  \multicolumn{14}{c}{\textit{Backbone Model: Qwen2.5-7B}} \\
  \midrule
  ReAct        & 2.85  & 1.94  & 0.58  & 4.00  & 2.10  & 4.04 & 0.52 & 4.34  & 10.67 & 9.32  & 9.23 & 1.36 & 6.79 \\
  Search-o1       & 18.52 & 16.73 & 5.42 & 16.00 & 16.10 & 3.92 & 4.11 & 15.87 & 32.45 & 22.18 & 25.37 & 1.28 & 19.82 \\
  R1-Searcher       & 47.21 & 44.85 & 20.63 & 44.00 & 43.02 & 3.91 & 11.00 & 45.12 & 63.28 & \textbf{49.05} & 54.07 & 1.08 & 50.06 \\
  Search-R1         & 46.64 & 45.60 & 19.97 & 45.90 & 43.19 & 3.77 & 11.45 & 44.35 & 63.65 & 48.70 & 53.96 & 1.05 & 51.39 \\
  OTC-PO       & 42.16 & 43.72 & 20.48 & 44.80 & 40.72 & \underline{1.56} & \underline{26.10} & 48.72 & 63.46 & 47.64 & 53.89 & \underline{1.00} & \underline{53.89} \\
  $\beta$-GRPO & 48.02 & \textbf{47.92} & 21.60 & 48.00 & 45.13 & 3.34 & 13.51 & 47.89 & 64.21 & 48.18 & 54.35 & 1.03 & 52.76 \\
  Offline AKBE & 47.38 & 46.81 & 20.95 & 46.40 & 44.22 & 2.68 & \textbf{16.50} & 47.15 & 63.92 & 48.05 & 54.08 & 1.01 & 53.54 \\
  \textbf{AKBE (Ours)} & \textbf{48.69} & 47.59 & \textbf{22.23} & \textbf{48.80} & \textbf{45.24} & 2.94 & 15.39 & \textbf{49.64} & \textbf{64.63} & 47.91 & \textbf{54.59} & 1.00 & \textbf{54.59} \\
  \bottomrule
  \end{tabular}
\end{table*}

\subsection{Experiment Settings}

\textbf{Datasets.} We evaluate AKBE on seven question answering benchmarks in a tool-augmented search setting. Following the setup of Search-R1\citep{jin2025search}, we deploy a lightweight search engine based on Wikipedia as the external tool environment. The benchmarks are organized into two categories: \textbf{Multi-Hop QA}, including HotpotQA \citep{yang2018hotpotqa}, 2WikiMultihopQA \citep{ho2020constructing}, MuSiQue \citep{trivedi2022musique}, and Bamboogle \citep{press2023measuring}, which require multi-step retrieval and reasoning; and \textbf{Single-Hop QA}, including Natural Questions (NQ) \citep{kwiatkowski2019natural}, TriviaQA \citep{joshi2017triviaqa}, and PopQA \citep{mallen2023llm_memorization}, which typically require a single retrieval. All benchmarks are evaluated using \textbf{Exact Match (EM)} as the primary metric. We additionally report \textbf{Tool Calls (TC)}, defined as the average number of tool calls per question, and \textbf{Tool Productivity (TP)}. $\text{TP} = \sum_{i=1}^{N} \mathbb{1}[R(y_i) = 1] \,/\, \sum_{i=1}^{N} \text{TC}(y_i)$, which measures accuracy per unit of tool usage.

\noindent \textbf{Baselines.} We compare AKBE against the following methods: (1) ReAct \citep{yao2023react}: a prompting-based approach, serving as the reference without RL training; (2) Search-o1 \citep{li2025search}: a framework that integrates an agentic search workflow into reasoning process; (3) R1-Searcher \citep{song2025r1} and (4) Search-R1 \citep{jin2025search}: two classic agentic RL frameworks that deploy GRPO for search enhancement; (5) OTC-PO \citep{wang2025acting}: a reward shaping method with a tool-productivity term to penalize redundant tool calls; (6) $\beta$-GRPO \citep{wu2025search}: a reward shaping method which introduces a confidence-based threshold to reduce uncertainty; and (7) Offline AKBE: an offline variant of AKBE that uses the same strategy of knowledge boundary-guided signal construction but generates the signal data from a fixed GRPO-trained checkpoint, serving as a direct comparison to validate the necessity of on-policy dynamic signal construction. Note that additional implementation specifics of baselines and AKBE are provided in Section~\ref{sec:implementation}.

\subsection{Main Results of AKBE}

We present the main results across two backbone models and seven benchmarks in Table~\ref{tab:main_results}. AKBE obtains the highest average EM score on both Multi-Hop and Single-Hop benchmarks while substantially reducing TC, yielding consistent TP improvements in most cases. On Qwen3-4B, AKBE improves EM by +1.85 on average across all seven benchmarks over its base method, while reducing TC by 18\%, yielding approximately a 25\% gain in tool productivity. The same effect holds on Qwen2.5-7B, confirming its generality across different model architectures and scales. In contrast, OTC-PO achieves the lowest TC across all settings (underlined in Table~\ref{tab:main_results}), but at a severe cost to accuracy, confirming that coarse-grained reward shaping incentivizes indiscriminate suppression of tool calls, leading to reward hacking. $\beta$-GRPO avoids EM collapse through its confidence threshold but provides limited TC reduction. AKBE achieves a strictly better balance: larger TC reduction than $\beta$-GRPO while simultaneously improving EM. 

Comparing AKBE with its offline variant (Offline AKBE) reveals the importance of on-policy signal construction. Offline AKBE consistently underperforms AKBE in EM score despite achieving even lower TC, reflecting overly aggressive ``reduce tool calls'' signals generated from the frozen trained policy. The knowledge boundary captured by offline data reflects the model's capability at a late training stage, which is overly optimistic for the weaker policy during early training. The resulting static boundary signals cannot align with the model's evolving knowledge state throughout training, leading to premature tool suppression and degraded accuracy. This validates our core claim that dynamic on-policy knowledge boundary tracking is essential for achieving the EM$\uparrow$TC$\downarrow$ balance.

\subsection{Analysis}

\subsubsection{Plug-and-Play Generalization}

Since AKBE enhances the model's knowledge boundary awareness through auxiliary supervisory signals rather than modifying the RL reward or optimization procedure, it is naturally orthogonal to the choice of base agentic RL algorithm and can serve as a plug-and-play module. To verify this, we integrate AKBE with four agentic RL algorithms: GRPO \citep{shao2024deepseekmathpushinglimitsmathematical}, DAPO \citep{yu2025dapo}, GSPO \citep{zheng2025groupsequencepolicyoptimization}, and AEPO \citep{dong2025agenticaepo}, each representing a distinct optimization strategy, such as dynamic sampling, sequence-level optimization, and entropy-driven exploration. 

As shown in Table~\ref{tab:plug_and_play}, AKBE consistently improves average EM and reduces TC across all four base algorithms. Notably, the improvements are consistent regardless of the base method's inherent nature: DAPO already achieves low TC ($2.61$) due to its dynamic sampling strategy for diverse trajectories, yet AKBE still further reduces it to $2.38$ while improving EM ($+0.36$ Avg.). For GSPO and AEPO, which exhibit higher base TC (3.23 and 3.08), AKBE delivers larger TC reductions ($-0.39$ and $-0.35$) alongside consistent EM gains ($+0.55$ and $+0.54$). The TP metric improves uniformly across all four pairings, with gains ranging from $+1.90$ to $+3.68$. These results confirm that AKBE acts as an efficiently orthogonal module. The boundary-guided training objective provides complementary learning signals that enhance tool call efficiency without interfering with the optimization dynamics of base RL algorithms.

\begin{table}[t]
  \centering
  \small
  \setlength{\tabcolsep}{2pt}
  \renewcommand{\arraystretch}{1.1}
  \caption{Plug-and-play results on Qwen3-4B Multi-Hop. AKBE consistently improves EM and reduces TC when combined with different agentic RL algorithms.}
  \label{tab:plug_and_play}
  \begin{tabular}{lccccccc}
  \toprule
  \textbf{Method} & \textbf{Hotpot} & \textbf{2Wiki} & \textbf{MuSi.} & \textbf{Bamb.} & \textbf{Avg.} & \textbf{TC}$\downarrow$ & \textbf{TP}$\uparrow$ \\
  \midrule
  GRPO        & 44.80 & 50.66 & 18.94 & 52.00 & 45.40 & 3.16 & 14.33 \\
  \textbf{+ AKBE} & \textbf{46.53} & \textbf{51.89} & \textbf{21.17} & \textbf{56.80} & \textbf{46.82} & \textbf{2.60} & \textbf{18.01} \\
  \midrule
  DAPO        & \textbf{45.95} & 51.81 & 21.68 & 51.20 & 46.65 & 2.61 & 17.85 \\
  \textbf{+ AKBE} & 45.83 & \textbf{52.35} & \textbf{22.51} & \textbf{52.80} & \textbf{47.01} & \textbf{2.38} & \textbf{19.75} \\
  \midrule
  GSPO        & 47.07 & 49.25 & 22.68 & \textbf{49.60} & 45.69  & 3.23 & 14.16 \\
  \textbf{+ AKBE} & \textbf{47.62} & \textbf{49.78} & \textbf{23.50} & 48.00 & \textbf{46.24} & \textbf{2.84} & \textbf{16.28} \\
  \midrule
  AEPO        & 46.36 & 51.78 & 23.47 & 52.00 & 46.96 & 3.08 & 15.25 \\
  \textbf{+ AKBE} & \textbf{46.89} & \textbf{52.26} & \textbf{24.24} & \textbf{54.40} & \textbf{47.50} & \textbf{2.73} & \textbf{17.40} \\
  \bottomrule
  \end{tabular}
\end{table}

\begin{table}[t]
  \centering
  \small
  \setlength{\tabcolsep}{6pt}
  \renewcommand{\arraystretch}{1.1}
  \caption{Ablation on trajectory signal categories (Qwen3-4B Multi-Hop). Removing each category reveals its unique contribution to AKBE.}
  \label{tab:ablation_category}
  \begin{tabular}{lccc}
  \toprule
  \textbf{Configuration} & \textbf{Avg. EM} & \textbf{TC}$\downarrow$ & \textbf{TP}$\uparrow$ \\
  \midrule
  GRPO & 45.40 & 3.16 & 14.33 \\
  \midrule
  \textbf{AKBE (Ours)} & \textbf{46.86} & \textbf{2.60} & \textbf{18.02} \\
  \quad w/o Tool-dependent & 43.56 & \underline{2.15} & \underline{20.26} \\
  \quad w/o Efficiency & 46.50 & 2.92 & 15.93 \\
  \quad w/o Hallucination & 46.55 & 2.58 & 18.04 \\
  \quad w/  Tool-dependent only & 46.02 & 2.85 & 16.15 \\
  \bottomrule
  \end{tabular}
\end{table}

\begin{figure}[t]
  \centering
  \includegraphics[width=1.0\linewidth]{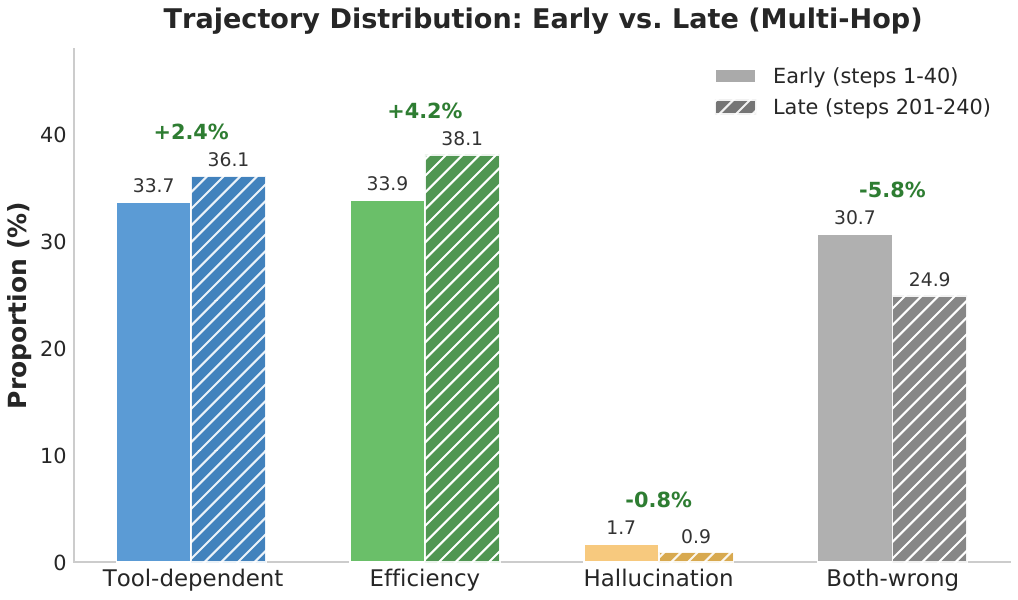}
  \caption{Trajectory category distribution at early vs.\ late training step on Qwen2.5-7B Multi-Hop. Both-wrong decreases substantially ($-$5.8\%), with the largest gain in Efficiency (+4.2\%), indicating knowledge internalization during AKBE training.}
  \label{fig:traj_dist}
\end{figure}

\subsubsection{Ablation Study on Trajectory Categories}

To understand the contribution of each signal category, we conduct ablation experiments by selectively removing individual categories from the knowledge boundary-guided training objective.

In Table~\ref{tab:ablation_category}, we find that removing Tool-dependent signals causes EM to drop \textit{below} GRPO significantly, despite achieving the lowest TC. The remaining Efficiency and Hallucination categories exclusively supervise toward no-tool trajectories, leading to over-suppression of necessary tool calls and degraded task accuracy. This confirms that Tool-dependent signals serve as a crucial protective mechanism that prevents the efficiency-oriented signals from over-suppressing necessary tool calls. Removing Efficiency signals yields a TC increase, identifying it as the primary force for eliminating redundant tool calls. Removing Hallucination signals results in a modest EM drop while TC remains comparable, validating that the Hallucination category corrects harmful tool-call paths where tool invocations override correct internal reasoning, contributing to the EM improvement. Notably, Tool-dependent alone already improves over GRPO, demonstrating that AKBE remains effective even on complex questions where no-tool rollouts mostly fail. Full AKBE outperforms all subsets, confirming the three categories are complementary: Tool-dependent teaches \textbf{\textit{when and how efficiently tools should be used}}, Efficiency teaches \textbf{\textit{when tools are unnecessary}}, and Hallucination teaches \textbf{\textit{when tools are harmful}}.

\subsubsection{Selection Strategy of the Coefficient $\lambda$}

\begin{figure*}[t]
  \centering
  \includegraphics[width=1.0\linewidth]{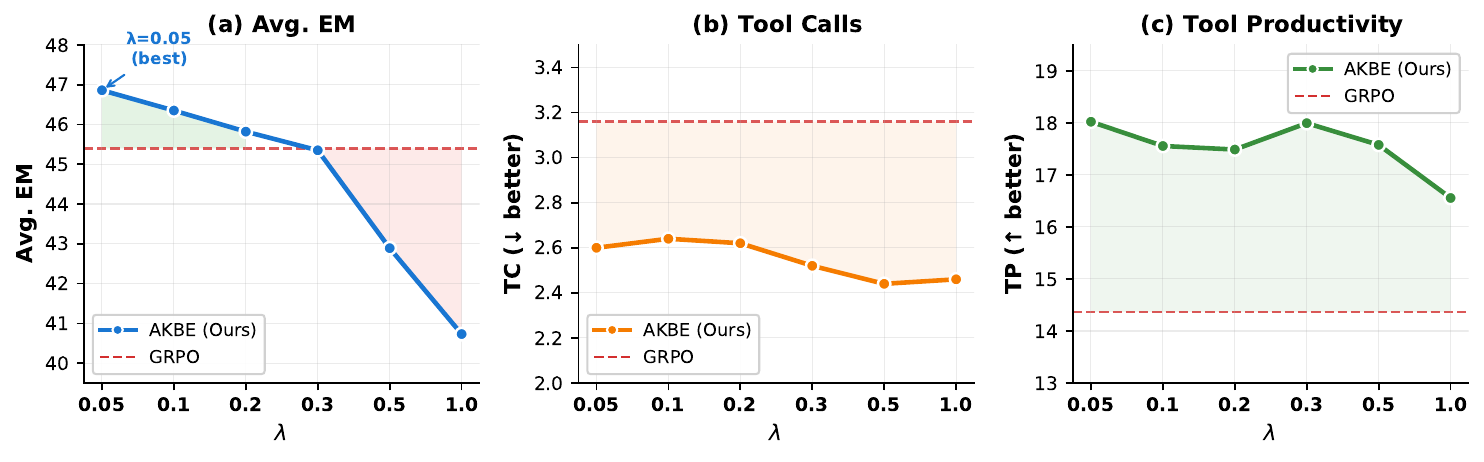}
  \caption{Effect of $\lambda$ on Qwen3-4B Multi-Hop. AKBE improves over GRPO for $\lambda \in [0.05, 0.2]$ (green region in (a)), with TC and TP consistently above GRPO across all values.}
  \label{fig:lambda}
\end{figure*}

We investigate the selection strategy of the coefficient $\lambda$ that balances the RL loss and the boundary-guided objective. Figure~\ref{fig:lambda} reports Avg.\ EM, TC, and TP on Qwen3-4B Multi-Hop as $\lambda$ varies in $\{0.05, 0.1, 0.2, 0.3, 0.5, 1.0\}$. AKBE consistently outperforms GRPO in EM across $\lambda \in [0.05, 0.2]$, with $\lambda = 0.05$ achieving the best balance. As $\lambda$ increases beyond 0.2, EM degrades sharply, indicating that an overly strong boundary-guided objective dominates the RL loss and leads to over-suppression of tool calls. Notably, TC and TP consistently outperform GRPO across all $\lambda$ values, indicating that AKBE reliably improves tool-use efficiency regardless of signal strength. The optimal $\lambda \approx 1/G_{wt}$ naturally balances the gradient contributions between the two objectives, as $\mathcal{L}_{\text{AKBE}}$ operates on at most one target trajectory per question while $\mathcal{L}_{\text{GRPO}}$ is computed over $G_{wt}$ rollouts. For this, we provide a detailed theoretical analysis in Appendix~\ref{sec:lambda_analysis}.

\begin{figure}[t]
  \centering
  \includegraphics[width=1.0\linewidth]{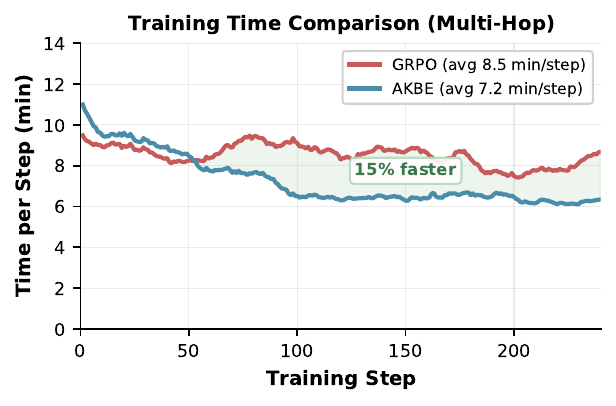}
  \caption{Per-step training time comparison on Qwen3-4B Multi-Hop. Despite additional no-tool rollouts, AKBE is 15\% faster on average due to efficient no-tool rollouts and reduced tool calls shortening overall time as training progresses.}
  \label{fig:overhead}
\end{figure}

\subsubsection{Trajectory Distribution During Training}

To examine how the knowledge boundary evolves during training, we compare the distribution of the trajectory categories between the early (steps 1--40) and late (steps 201--240) training phases on Qwen2.5-7B Multi-Hop in Figure~\ref{fig:traj_dist}. 

The most obvious change is a substantial decrease in Both-wrong proportion, indicating that agentic RL training progressively enables the model to solve previously intractable questions. Crucially, the Efficiency category shows the largest increase, demonstrating that AKBE successfully promotes knowledge internalization, where the model increasingly learns to answer questions using its parametric knowledge. Meanwhile, Hallucination decreases notably, confirming that Hallucination signals effectively correct harmful tool-call paths during training. These shifts validate two key aspects of our design: (1) the knowledge boundary is non-static during training, justifying on-policy signal construction over static offline approaches, and (2) AKBE's boundary-guided objective and the RL objective work synergistically, where RL strengthens the tool-augmented reasoning capability of model while AKBE delivers knowledge boundary-guided efficiency signals that guide the model to maximize the utilization of its knowledge boundary, achieving efficient reasoning paths with minimal redundant tool calls.

\subsubsection{Computational Overhead}

A natural concern is whether the additional no-tool rollouts in AKBE introduce prohibitive computational overhead. Figure~\ref{fig:overhead} compares the time consumption per training step between GRPO and AKBE on Qwen3-4B Multi-Hop. Surprisingly, AKBE is on average 15\% \textit{faster} than GRPO, despite performing additional $G_{nt} = 8$ no-tool rollouts per batch. This result arises from two factors: (1) no-tool rollouts complete substantially faster than with-tool rollouts as they involve no tool interaction or environment latency, and (2) as AKBE progressively reduces tool calls during training, the with-tool rollouts themselves become shorter with fewer tool calls, leading to accelerating step times in later training stages. This demonstrates that AKBE introduces little computational overhead, which pays for itself through the efficiency gains it induces in most cases. We further provide a detailed per-step comparison of tool call counts and response lengths in Appendix~\ref{sec:efficiency_details}.

\section{Conclusion}

In this paper, we presented AKBE, a simple but effective method that dynamically probes the model's intrinsic knowledge boundary through on-policy dual-path rollouts during agentic RL training. By constructing knowledge boundary-guided supervisory signals, AKBE eliminates redundant tool calls while preserving necessary ones, and guides the model toward efficient tool-call patterns. Unlike reward shaping approaches that suffer from reward hacking, AKBE provides more fine-grained guidance at the instance level without modifying the RL objective, enabling simultaneous improvement in task accuracy and tool-call efficiency. Experiments across seven QA benchmarks and two backbone models validate its effectiveness, demonstrating that explicit on-policy knowledge boundary modeling is a promising and general strategy for efficient agentic reinforcement learning.

\section{Limitations}

Although AKBE achieves faster average training time than GRPO due to reduced tool calls in later stages, the additional no-tool rollouts do introduce extra computational cost in the early training phase when tool calls have not yet decreased. Future work could explore more efficient rollout strategies, such as adaptive sampling that selectively performs no-tool rollouts only for questions likely to be within the knowledge boundary. Besides, the coefficient $\lambda$ is fixed throughout training, while the optimal balance between $\mathcal{L}_{\text{RL}}$ and $\mathcal{L}_{\text{AKBE}}$ may vary across training stages and task difficulties. An adaptive $\lambda$ that adjusts per step based on the current trajectory distribution or task complexity could further improve performance.

\bibliography{custom}

\newpage

\appendix

\section{Implementation Details}
\label{sec:implementation}

\subsection{Reward Design}

Our training pipeline employs a binary outcome reward that combines answer correctness with a structural format requirement. The correctness signal follows the reward formulation of Search-R1 \citep{jin2025search}, using Exact Match as the primary evaluation criterion.

\paragraph{Exact Match Reward.} Given the final answer $\hat{y}$ extracted from the agent's trajectory and the ground-truth answer $y^*$, the EM reward is defined as:
\begin{equation}
    r_{\text{EM}}(\hat{y}, y^*) = \begin{cases} 1, & \text{if } \hat{y} = y^* \\ 0, & \text{otherwise} \end{cases}
\end{equation}
This strict binary formulation eliminates the ambiguity of partial-credit scoring and drives the policy toward fully correct answers, providing a clear optimization signal for agentic RL.

\paragraph{Format Constraint.} In addition to correctness, each trajectory must satisfy a structural validity requirement. The response must contain both a reasoning trace wrapped by \texttt{<think>...</think>} tags and a final answer wrapped by \texttt{<answer>...</answer>} tags, with the answer further enclosed in \texttt{\textbackslash boxed\{\}}. The format indicator is:
\begin{equation}
    I_{\text{format}} = \begin{cases} 1, & \text{if all required tags are present} \\ 0, & \text{otherwise} \end{cases}
\end{equation}
Responses violating this schema receive no credit regardless of answer correctness, ensuring reliable tool-call parsing and final-answer extraction.

\paragraph{Final Reward.} The overall reward combines both components:
\begin{equation}
    r = \begin{cases} r_{\text{EM}}(\hat{y}, y^*), & \text{if } I_{\text{format}} = 1 \\ -1, & \text{otherwise} \end{cases}
\end{equation}
A trajectory earns the maximal reward of 1 only when it satisfies the format requirement and delivers an exactly correct answer; format violations are explicitly penalized with $r = -1$.

\subsection{Prompt Template}

\begin{figure*}[t]
\centering
\begin{tcolorbox}[
    colback=white,
    colframe=black,
    arc=10pt,
    boxrule=1.5pt,
    left=10pt,
    right=10pt,
    top=5pt,
    bottom=5pt,
    toptitle=3pt,
    bottomtitle=3pt,
    title={\centering\textcolor{black}{\normalsize\bfseries Prompt Template for With-tool Rollout}},
    colbacktitle=white,
    coltitle=white,
    fonttitle=\bfseries\Large
]

You are a helpful assistant that can solve the given question step by step with the help of the wikipedia search tool. Given a question, you need to first think about the reasoning process in the mind and then provide the answer. During thinking, you can invoke the wikipedia search tool to search for fact information about specific topics if needed. You can search as many times as your want. The reasoning process and answer are enclosed within \textcolor{myblue}{<think> </think>} and \textcolor{mypurple}{<answer> </answer>} tags respectively, and the search query and result are enclosed within \textcolor{myred}{<search> </search>} and \textcolor{mygreen}{<result> </result>} tags respectively. For example, \textcolor{myblue}{<think>} This is the reasoning process. \textcolor{myblue}{</think>} \textcolor{myred}{<search>} search query here \textcolor{myred}{</search>} \textcolor{mygreen}{<result>} search result here \textcolor{mygreen}{</result>} \textcolor{myblue}{<think>} This is the reasoning process. \textcolor{myblue}{</think>} \textcolor{mypurple}{<answer>} The final answer is [ \textbf{\textbackslash{}boxed\{answer here\}} ] \textcolor{mypurple}{</answer>}. In the last part of the answer, the final exact answer is enclosed within \textbf{\textbackslash{}boxed\{\}} with latex format. Question:

\end{tcolorbox}
\caption{The prompt template for with-tool rollout in our experiment setting.}
\label{with-tool prompt template}
\end{figure*}

\begin{figure*}[t]
\centering
\begin{tcolorbox}[
    colback=white,
    colframe=black,
    arc=10pt,
    boxrule=1.5pt,
    left=10pt,
    right=10pt,
    top=5pt,
    bottom=5pt,
    toptitle=3pt,
    bottomtitle=3pt,
    title={\centering\textcolor{black}{\normalsize\bfseries Prompt Template for No-tool Rollout}},
    colbacktitle=white,
    coltitle=white,
    fonttitle=\bfseries\Large
]

You are a helpful assistant that can solve the given question step by step without external tool. Given a question, you need to first think about the reasoning process in the mind and then provide the answer. The reasoning process and answer are enclosed within \textcolor{myblue}{<think> </think>} and \textcolor{mypurple}{<answer> </answer>} tags respectively. For example, \textcolor{myblue}{<think>} This is the reasoning process. \textcolor{myblue}{</think>} \textcolor{mypurple}{<answer>} The final answer is [ \textbf{\textbackslash{}boxed\{answer here\}} ] \textcolor{mypurple}{</answer>}. In the last part of the answer, the final exact answer is enclosed within \textbf{\textbackslash{}boxed\{\}} with latex format. Question:

\end{tcolorbox}
\caption{The prompt template for no-tool rollout in our experiment setting.}
\label{no-tool prompt template}
\end{figure*}

AKBE requires two prompt templates for its dual-path rollouts, as shown in Figure~\ref{with-tool prompt template} and Figure~\ref{no-tool prompt template}.

\paragraph{With-tool Prompt.} The with-tool template (Figure~\ref{with-tool prompt template}) follows the tag-based format adopted in prior agentic RL work \citep{jin2025search, dong2025agenticaepo, zong2026at2po}. Each rollout is structured into semantically distinct regions delimited by dedicated tag pairs: reasoning steps are verbalized within \textcolor{myblue}{\textit{<think></think>}}, retrieval queries are issued via \textcolor{myred}{\textit{<search></search>}}, environment observations are injected within \textcolor{mygreen}{\textit{<result></result>}}, and the final prediction is emitted within \textcolor{mypurple}{\textit{<answer></answer>}} with the canonical answer enclosed in \textbf{\textbackslash{}boxed\{\}} for Exact Match extraction.

\paragraph{No-tool Prompt.} The no-tool template (Figure~\ref{no-tool prompt template}) removes all tool-related instructions and tags (\textcolor{myred}{\textit{<search>}} and \textcolor{mygreen}{\textit{<result>}}), retaining only the reasoning (\textcolor{myblue}{\textit{<think></think>}}) and answer (\textcolor{mypurple}{\textit{<answer></answer>}}) components. This forces the model to generate answers solely from its parametric knowledge, enabling AKBE to probe the knowledge boundary by comparing correctness across the two paths. 

\subsection{Datasets}

We conduct experiments on two categories of widely-used question answering benchmarks to evaluate the effectiveness of our proposed AKBE.

\paragraph{Multi-Hop QA.} This category evaluates multi-turn tool use and compositional reasoning, where correct answers cannot be obtained from a single retrieved passage. \textbf{HotpotQA} \citep{yang2018hotpotqa} is a large-scale Wikipedia-derived benchmark with supporting-fact annotations, serving as a widely used testbed for multi-hop question answering. \textbf{2WikiMultiHopQA} \citep{ho2020constructing} combines Wikipedia passages with Wikidata triples, producing questions that require explicit multi-hop entity reasoning. \textbf{MuSiQue} \citep{trivedi2022musique} contains approximately 25k questions spanning 2--4 reasoning hops, synthesized through controlled composition of single-hop primitives to probe fine-grained reasoning depth. \textbf{Bamboogle} \citep{press2023measuring} offers a small but adversarial set of compositional queries, serving as a robustness probe for agentic RL policies.

\paragraph{Single-Hop QA.} This category verifies performance on single-step retrieval tasks. \textbf{Natural Questions (NQ)} \citep{kwiatkowski2019natural} aggregates real user queries answered from Wikipedia and serves as a standard benchmark for retrieval-augmented generation. \textbf{TriviaQA} \citep{joshi2017triviaqa} features substantial lexical and syntactic divergence between questions and supporting evidence, testing robustness to surface variation. \textbf{PopQA} \citep{mallen2023llm_memorization} is an entity-centric benchmark designed to separate the contribution of external retrieval from parametric memorization, making it a natural diagnostic for whether the policy genuinely leverages the search tool versus relying on memorized facts.

\subsection{AKBE Settings}

For implementation details of our AKBE, we use a training batch size of $64$, a mini-batch size of $8$, and a maximum response length of $6192$. During rollout, we use a rollout size of $16$ for with-tool rollout and a rollout size of $8$ for no-tool rollout, with the maximum tool usage set to $6$. The clipping thresholds for the AKBE objective are set to $0.2$ (the same as GRPO). Following prior work \citep{yu2025dapo, zheng2025groupsequencepolicyoptimization}, we remove the KL regularization term ($\beta = 0$) to allow the policy to explore diverse rollout strategies. The AKBE coefficient is set to $\lambda = 0.05$ for both Multi-Hop and Single-Hop settings, which empirically balances the gradient contributions between $\mathcal{L}_{\text{GRPO}}$ and $\mathcal{L}_{\text{AKBE}}$ (see \S\ref{sec:lambda_analysis} for detailed analysis).

\subsection{Baseline Settings}

\begin{table}[t]
  \centering
  \small
  \setlength\tabcolsep{8pt}
  \renewcommand\arraystretch{1.2}
  \caption{Shared hyperparameters used by the agentic RL baselines in our experiment.}
  \begin{tabular}{lcccccccc}
  \hline
  Config & Value  \\
  \hline
  optimizer & AdamW  \\
  learning rate & 1e-6  \\
  clip\_ratio & 0.2 \\
  training batch size & 64 \\
  PPO mini batch size & 8 \\
  rollout\_n & 16 \\
  max prompt length & 2000 \\
  max response length & 6192  \\
  max tool-call turns & 6 \\
  reward metrics & EM  \\
  retriever & local wiki \\
  top-K retrieval passages & 3 \\
  \hline
  \end{tabular}
  
  \label{tab:hyperparameters}
  \end{table}

Table~\ref{tab:hyperparameters} summarizes the shared hyperparameters used across all RL-based baselines. For method-specific configurations, we follow the settings reported in the respective original papers. All baselines are trained without any additional SFT phase. We select and report results from the checkpoint achieving the highest average EM across all evaluation benchmarks.

\subsection{Search Tool Environment}

Our search tool environment follows the setup of Search-R1 \citep{jin2025search}. We use a Wikipedia snapshot as the retrieval corpus with e5-base-v2 \citep{wang2022text} as the dense retriever. The knowledge base contains approximately 21M Wikipedia entries, providing broad factual coverage for both single-hop and multi-hop queries. At each turn where the policy emits a retrieval action, the search engine scores candidate passages against the query and returns the top-3 most relevant entries, which are injected into the context as tool observations for reasoning.

\subsection{Hardware and Artifacts}

All training and evaluation experiments are conducted on a single node with 8$\times$ NVIDIA H20 GPUs. We adopt two publicly available checkpoints as backbone policies: Qwen3-4B \citep{yang2025qwen3technicalreport} and Qwen2.5-7B \citep{qwen2025qwen25technicalreport}, selected for their strong reasoning capabilities and demonstrated compatibility with agentic post-training. Our training infrastructure is built on the VeRL framework \citep{sheng2024hybridflow}, a hybrid-controller RL system whose modular rollout interface supports the multi-turn, tool-interactive rollout schedule required by AKBE's dual-path design.

\section{Theoretical Analysis for Coefficient $\lambda$}
\label{sec:lambda_analysis}

We provide a theoretical analysis for the optimal value of $\lambda$ from the perspective of gradient contribution balancing between $\mathcal{L}_{\text{GRPO}}$ and $\mathcal{L}_{\text{AKBE}}$.

\paragraph{Setup.} Consider a training batch of $B$ questions. For each question $q$, the GRPO objective is computed over $G_{wt}$ with-tool rollout trajectories, while the AKBE objective selects at most one target trajectory $y_q^*$. The total training loss is:
\begin{equation}
    \mathcal{L}_{\text{total}} = \mathcal{L}_{\text{GRPO}} + \lambda \cdot \mathcal{L}_{\text{AKBE}}
\end{equation}

\paragraph{Gradient Analysis.} The per-question gradient contribution from the GRPO loss involves $G_{wt}$ trajectories:
\begin{equation}
    \nabla_\theta \mathcal{L}_{\text{GRPO}}^{(q)} = \frac{1}{G_{wt}} \sum_{i=1}^{G_{wt}} \hat{A}_i \cdot \nabla_\theta \log \pi_\theta(y_i \mid q)
\end{equation}
where $\hat{A}_i$ is the group-relative advantage. Under the assumption that each trajectory contributes approximately equal gradient magnitude $\|\nabla_\theta \log \pi_\theta(y_i \mid q)\| \approx g$, the expected per-question gradient norm is:
\begin{equation}
    \mathbb{E}\left[\|\nabla_\theta \mathcal{L}_{\text{GRPO}}^{(q)}\|\right] \approx \sigma_A \cdot g
\end{equation}
where $\sigma_A$ is the standard deviation of the advantage estimates within the group.

For the AKBE loss, each question contributes at most one target trajectory:
\begin{equation}
    \nabla_\theta \mathcal{L}_{\text{AKBE}}^{(q)} = -\nabla_\theta \log \pi_\theta(y_q^* \mid q)
\end{equation}
with gradient norm approximately $g$. However, not all questions produce a signal. Let $p_s$ denote the proportion of questions with constructible signals ($p_s = |\mathcal{S}| / B$). The effective per-question gradient contribution from AKBE (averaged over the batch) is:
\begin{equation}
    \mathbb{E}\left[\|\nabla_\theta \mathcal{L}_{\text{AKBE}}^{(q)}\|\right] \approx p_s \cdot g
\end{equation}

\begin{figure*}[t]
  \centering
  \includegraphics[width=0.8\linewidth]{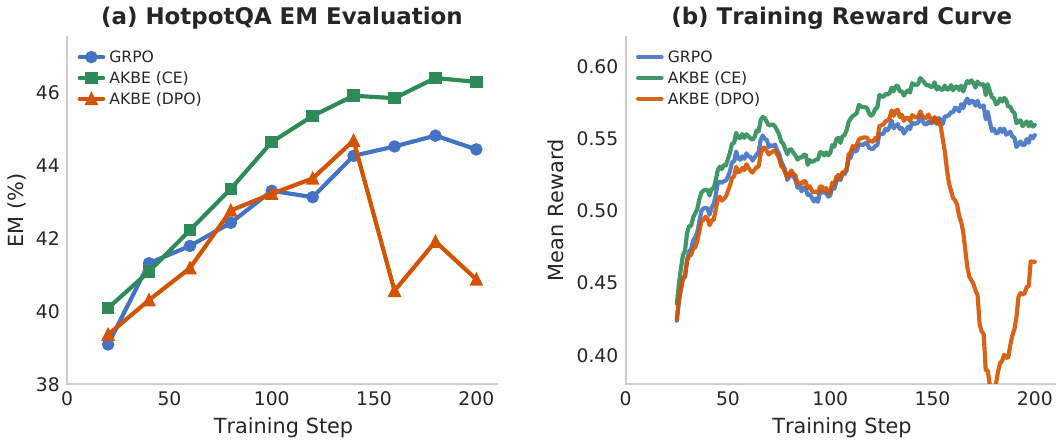}
  \caption{Comparison of signal integration methods on Qwen3-4B Multi-Hop. \textit{Left}: EM evaluation over training steps. \textit{Right}: Training reward curves. DPO-based integration shows initial promise but collapses in later training.}
  \label{fig:dpo_analysis}
\end{figure*}

\paragraph{Balancing Condition.} For the two objectives to contribute comparably to the overall parameter update, we require:
\begin{equation}
    \|\nabla_\theta \mathcal{L}_{\text{GRPO}}\| \approx \lambda \cdot \|\nabla_\theta \mathcal{L}_{\text{AKBE}}\|
\end{equation}
Substituting the per-question estimates and noting that the GRPO loss normalizes over $G_{wt}$ trajectories while AKBE operates on single trajectories:
\begin{align}
    \sigma_A \cdot g &\approx \lambda \cdot p_s \cdot g \notag \\
    \lambda &\approx \frac{\sigma_A}{p_s}
\end{align}

\paragraph{Practical Estimate.} In our setting, the advantage standard deviation under binary rewards ($R \in \{0, 1\}$) with group-relative normalization is $\sigma_A \approx 1$ by construction. The signal proportion $p_s$ is typically high (around 70--80\% of questions produce at least one correct trajectory in either path). However, the key scaling factor is the ratio of trajectory counts: $\mathcal{L}_{\text{GRPO}}$ aggregates gradients from $G_{wt}$ trajectories per question (each weighted by $1/G_{wt}$), while $\mathcal{L}_{\text{AKBE}}$ uses exactly one trajectory at full weight. To prevent the single AKBE trajectory from dominating the $G_{wt}$ RL trajectories, $\lambda$ should scale as:
\begin{equation}
    \lambda \approx \frac{1}{G_{wt}}
\end{equation}

With $G_{wt} = 16$ in our experiments, this yields $\lambda \approx 0.0625$, closely matching our empirical optimum of $\lambda = 0.05$. Considering dynamic factors such as task difficulty and signal proportion variability, we find that $\lambda \in [0.05, 0.10]$ constitutes a reasonable range in our experimental setting.

\section{Cross-Entropy vs. DPO for Signal Integration}

A natural concern is whether the boundary-guided signals could be integrated via preference optimization (e.g., DPO) rather than cross-entropy. Since AKBE's signal construction identifies preferred trajectories for each question, one could additionally select rejected trajectories and apply a DPO-style objective:
\begin{align}
    \mathcal{L}_{\text{DPO}} = -\log \sigma\Big(\beta \log \frac{\pi_\theta(y_w \mid q)}{\pi_{\text{ref}}(y_w \mid q)} &\notag \\
    - \beta \log \frac{\pi_\theta(y_l \mid q)}{\pi_{\text{ref}}(y_l \mid q)}&\Big)
\end{align}
where $y_w$ and $y_l$ denote the preferred and rejected trajectories respectively.

\paragraph{Signal Construction Difference.} The AKBE based on cross-entropy formulation retains only positive signals, supervising the model toward the selected target trajectory without explicit penalization of alternatives. In contrast, the DPO-based variant additionally selects rejected trajectories from incorrect or inefficient rollouts, specifically choosing the trajectory with the highest tool-call count (i.e., maximum divergence from the preferred pattern) as the negative signal. This creates explicit preference pairs that contrast efficient and inefficient tool-use behaviors.

\paragraph{Experimental Comparison.} We compare three configurations on Qwen3-4B Multi-Hop: standard GRPO, AKBE with cross-entropy (our method), and AKBE with DPO-based integration. Figure~\ref{fig:dpo_analysis} presents the EM evaluation and training reward dynamics. $\lambda = 0.05$ is applied to both experiments.

\paragraph{Results and Analysis.} As shown in Figure~\ref{fig:dpo_analysis}, the DPO-based variant initially shows promising performance, even surpassing GRPO during the mid-training phase (steps 80--140). However, it subsequently undergoes a sudden collapse, with both EM and training reward dropping sharply after step 150. In contrast, AKBE (CE based) maintains stable and monotonic improvement throughout training.

We attribute this instability to the inherent similarity between preferred and rejected trajectory patterns. Both positive and negative signals are trajectories that involve reasoning and tool calls; they differ only in the tool-call strategies or final correctness. As DPO explicitly reduces the probability of rejected trajectories, the model gradually learns to penalize the shared tool-call patterns present in both preferred and rejected samples, rather than learning the fine-grained distinction between efficient and inefficient tool use. This causes a progressive over-suppression of tool-call behavior that eventually leads to training collapse. Cross-entropy avoids this failure mode by providing positive-only supervision: it teaches the model what efficient patterns looks like without explicitly penalizing alternatives, resulting in a more stable and targeted optimization signal for knowledge boundary enhancement.

\section{Reliability of Knowledge Boundary Estimation}
\label{sec:kb_reliability}

\begin{figure}[t]
  \centering
  \includegraphics[width=1.0\linewidth]{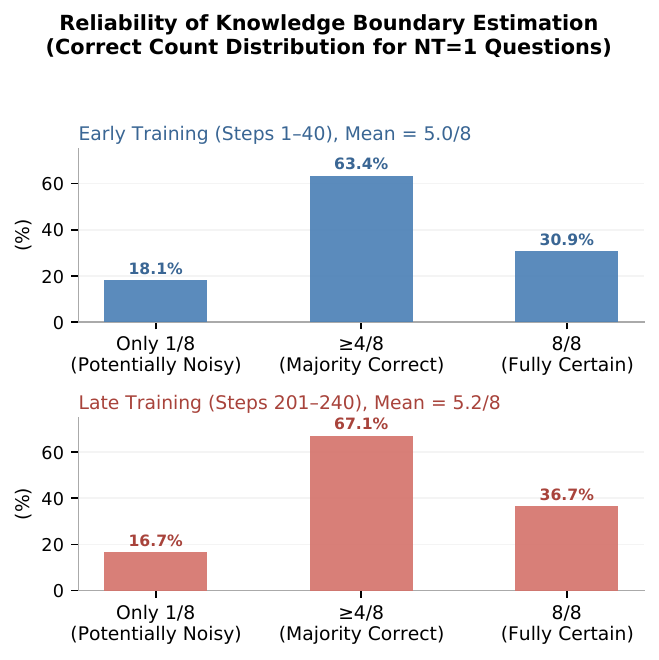}
  \caption{Distribution of correct no-tool trajectory counts for which classified as NT=1 ($G_{nt}=8$) on Qwen3-4B Multi-Hop. The majority of NT=1 questions have $\geq$4/8 correct rollouts, indicating reliable knowledge boundary estimation.}
  \label{fig:kb_reliability}
\end{figure}

We discuss a potential concern of whether the knowledge boundary estimation based on ``at least one correct'' among no-tool rollouts ($G_{nt}=8$ in our experiment settings) is reliable or whether it is dominated by questions where the model merely guesses correctly by chance.

Specifically, we analyze the distribution of correct rollout counts for all questions classified as NT=1 across training. As shown in Figure~\ref{fig:kb_reliability}, NT=1 questions have 5.0/8 correct rollouts in early training and 5.2/8 in late training on average. Only 18.1\% (early) and 16.7\% (late) of NT=1 questions have exactly 1/8 correct, while the majority (63.4\% early, 67.1\% late) achieve $\geq$4/8 correct rollouts. Notably, 30.9\% (early) to 36.7\% (late) of questions achieve a perfect 8/8, indicating that the model fully and reliably leverages the parametric knowledge.

These results demonstrate that the knowledge boundary estimation is in high confidence rather than noise-driven. Furthermore, the improvement from early to late training confirms that AKBE's on-policy design progressively strengthens the reliability of boundary estimation as training progresses. For the minority of potentially noisy cases (1/8 correct), the small coefficient $\lambda=0.05$ ensures that these weak signals cannot override the dominant RL objective, and the on-policy mechanism provides an additional indemnification.

\section{Supplementary Overhead Analysis}
\label{sec:efficiency_details}

\begin{figure*}[t]
  \centering
  \includegraphics[width=0.85\linewidth]{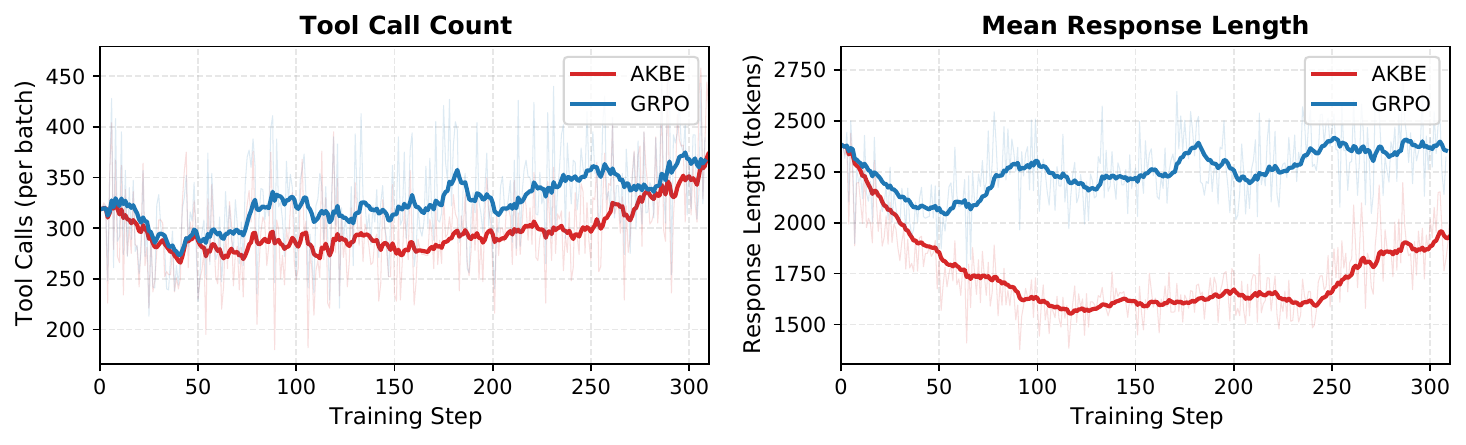}
  \caption{Per-step comparison of tool call counts (\textit{Left}) and mean response length (\textit{Right}) between GRPO and AKBE on Qwen3-4B Multi-Hop. AKBE consistently yields fewer tool calls and shorter responses during training.}
  \label{fig:efficiency_details}
\end{figure*}

To complement the per-step overhead comparison in the main text (\S5.3.5), we provide a supplementary analysis of the two underlying factors that drive AKBE's computational efficiency: tool call frequency and response length.

\paragraph{Tool Call Count.} Figure~\ref{fig:efficiency_details} (Left) shows the total number of search calls per training batch across training steps. Both methods start at a similar level ($\sim$320 calls per batch), but their trajectories diverge as training progresses. GRPO's tool call count increases steadily, rising from approximately 320 to over 370 by step 300, reflecting the well-known tendency of RL-trained agents to escalate tool use when rewarded only for correctness. In contrast, AKBE maintains a relatively stable tool call count around 290--300 throughout training, with a slight decrease in the middle stages (steps 50--150). On average, AKBE reduces tool calls by 8.4\% relative to GRPO across the entire training process.

\paragraph{Response Length.} Figure~\ref{fig:efficiency_details} (Right) reveals significant difference in mean response length. GRPO exhibits a strong upward trend, with average response length growing from $\sim$2,300 tokens to over 2,500 tokens by step 300. This increase correlates with more tool calls generating longer multi-turn interaction sequences. AKBE, by contrast, shows a decreasing trend in the early-to-middle training phase (steps 1--100), stabilizing around 1,600--1,700 tokens thereafter. The overall response length reduction is 23.0\%, directly translating to lower inference latency and computational cost.

These trends confirm that AKBE's knowledge boundary-guided signals produce a compounding efficiency effect: by teaching the model to avoid unnecessary tool calls early in training, subsequent rollouts become inherently shorter, which in turn accelerates both training and inference.

\section{Case Study}

We present representative examples from three AKBE signal categories during training (Qwen3-4B Multi-Hop). These cases illustrate how the dual-path comparison reveals the model's knowledge boundary and guides signal construction. For each case, we show the complete reasoning trajectories from both paths, the signal category classification, and which trajectory is selected as the supervision target.

\paragraph{Analysis of Case \#1 (Efficiency).} This case demonstrates the \textit{efficiency} signal category, where the model possesses sufficient parametric knowledge to answer correctly without external retrieval. In the with-tool path, the model explicitly states ``I'm not immediately familiar with Donker Mag'' and initiates a search, despite the no-tool path revealing that it can readily recall the album's association with Die Antwoord from its internal knowledge. This discrepancy exposes a Level-1 knowledge boundary violation: the model defaults to tool use even when unnecessary. By selecting the no-tool trajectory as the supervision target, AKBE teaches the model to trust its parametric memory for well-known facts, reducing redundant tool calls and improving inference efficiency.

\paragraph{Analysis of Case \#2 (Tool-dependent).} This case illustrates the \textit{tool-dependent} category, where external retrieval is genuinely necessary. The no-tool path exposes a clear knowledge gap: the model hallucinates a plausible but incorrect answer (``Warren Buffett''), cycling through multiple guesses without arriving at the correct one (Seth Klarman). In contrast, the with-tool path formulates a targeted query (``Baupost Group founder'') that immediately retrieves the relevant passage, enabling correct reasoning. Here, the with-tool trajectory with minimum tool calls (TC=1) is selected as the target. This signal reinforces Level-2 boundary awareness: for questions beyond parametric knowledge, the model should learn efficient retrieval patterns that resolve the query in a single tool call.

\paragraph{Analysis of Case \#3 (Hallucination).} This case reveals a subtle failure mode: \textit{retrieval-induced hallucination}. The model searches for ``Arline Burks Gant death date'' but the retriever returns irrelevant passages about ``Barbara Stoddard Burks.'' Rather than recognizing the retrieval failure, the with-tool path incorrectly conflates the two individuals based on the shared surname, arriving at a wrong answer through flawed reasoning (``maybe Arline Burks Gant is the same as Barbara Stoddard Burks?''). Meanwhile, the no-tool path correctly recalls that Maurice Pialat died in 2003 and Arline Burks Gant lived until 2011. This case demonstrates that tool use can be actively harmful when retrieval results are noisy or off-target. By selecting the no-tool trajectory as the target, AKBE guides the model to avoid over-reliance on retrieval for queries where parametric knowledge is more reliable than noisy search results.

These three cases collectively illustrate how AKBE's dual-path comparison mechanism adaptively identifies the appropriate behavior at the knowledge boundary. The signal construction does not uniformly favor either path; instead, it selects the trajectory that reflects the most appropriate tool-use decision for each specific query. This fine-grained, per-question supervision enables the model to develop nuanced knowledge boundary awareness across diverse scenarios.

\begin{table*}[!t]
    \centering
    \caption{Case \#1: \textbf{Efficiency} ($WT$=\checkmark, $NT$=\checkmark). The model already knows the answer from parametric knowledge; the tool call is redundant. The no-tool trajectory is selected as the target signal.}
    \fontsize{8.5pt}{10.5pt}\selectfont
    \begin{tabular}{@{}p{\textwidth}@{}}
    \toprule
        \textbf{Question:} Donker Mag is an album by what African hip hop group? \quad \textbf{Ground Truth:} Die Antwoord \\
    \midrule
        \rowcolor{blue!5}\textbf{With-tool path} (TC=1, EM=1.0) \\
        \blue{<think>} Okay, the user is asking about the album ``Donker Mag'' and wants to know which African hip hop group it belongs to. First, I need to figure out who the artists are associated with this album. I'm not immediately familiar with ``Donker Mag,'' so I should look it up. \blue{</think>}
        \\\red{<search>} Donker Mag album \red{</search>}
        \\\green{<result>}
        \\Page 1: ``Die Antwoord'' However the album was met with average to poor reviews by many critics. ``Ten\$Ion'' was followed up with a non-album single, ``XP N\$IV \$H1T''. Following the release of ``XP N\$IV \$H1T'', Die Antwoord set out on a brief tour across Europe from June to July. Soon after they began touring, snippets for a new single called ``Cookie Thumper!'', which was to be accompanied by a music video, were released. Around the release of their new single, Die Antwoord also announced the title of their third album, ``Donker Mag'', which was released on 3 June 2014. The video for ``Cookie Thumper!'' was released on 27 May 2014 as the album's lead single.
        \\Page 2: ``Magnet (magazine)'' Magnet is a music magazine that generally focuses on alternative, independent, or out-of-the-mainstream bands. The magazine is published four times a year, and is independently owned and edited by Eric T. Miller. Music magazines with a similar focus in the 1990s era included ``Option'', ``Raygun'', and ``Alternative Press''. The first issue of ``Magnet'' came out in mid-1993.
        \\Page 3: ``Green Monkey Records'' (CDR GM1028) The OF - Escape Goat (CD GM1029) Jim McIver - Sunlight Reaches (CD GM1030) Tom Dyer's New Pagan Gods - History Of Northwest Rock Vol. 1, 1959-1968 (CD GM1031) Fur For Fairies (CD GM1032) The Fuzz - Best Kept Secret...
        \\\green{</result>}
        \\\blue{<think>} Hmm, the first search result mentions ``Die Antwoord'' in relation to the album ``Donker Mag.'' Die Antwoord is a South African hip hop group. The other results don't seem relevant. So, the answer should be Die Antwoord. \blue{</think>}
        \\\purple{<answer>} The final answer is \textbackslash boxed\{Die Antwoord\}. \purple{</answer>} \\
    \midrule
        \rowcolor{green!5}\textbf{No-tool path} (TC=0, EM=1.0) $\leftarrow$ \textit{Selected as target} \\
        \blue{<think>} Okay, the user is asking about the album ``Donker Mag'' and which African hip hop group it belongs to. First, I need to recall if I know any African hip hop groups that have an album with that title. ``Donker Mag'' sounds Dutch. Let me think... I recall that Die Antwoord is a South African hip hop group known for performing in Afrikaans and English. ``Donker Mag'' means ``Dark Force'' in Afrikaans/Dutch. Yes, I'm fairly confident that ``Donker Mag'' is the third studio album by Die Antwoord, released on 3 June 2014. The group consists of Ninja and Yolandi Visser, and they are from Cape Town, South Africa. \blue{</think>}
        \\\purple{<answer>} The final answer is \textbackslash boxed\{Die Antwoord\}. \purple{</answer>} \\
    \bottomrule
    \end{tabular}
\end{table*}

\begin{table*}[!t]
    \centering
    \caption{Case \#2: \textbf{Tool-dependent} ($WT$=\checkmark, $NT$=\ding{55}). The model lacks parametric knowledge about this specific fact. The minimum tool-call with-tool trajectory is selected as the target signal.}
    \fontsize{8.5pt}{10.5pt}\selectfont
    \begin{tabular}{@{}p{\textwidth}@{}}
    \toprule
        \textbf{Question:} The Baupost Group asked which American investor, hedge fund manager, and philanthropist to run the hedge fund? \quad \textbf{Ground Truth:} Seth Klarman \\
    \midrule
        \rowcolor{blue!5}\textbf{With-tool path} (TC=1, EM=1.0) $\leftarrow$ \textit{Selected as target (min TC)} \\
        \blue{<think>} Okay, let's see. The question is about the Baupost Group asking which American investor, hedge fund manager, and philanthropist to run the hedge fund. I need to figure out the answer. First, I should recall who the Baupost Group is. Baupost is a hedge fund firm, right? They might have had a notable founder or a prominent figure associated with them. The question mentions that they asked a specific person to run the fund. So, I need to find the name of that individual. Maybe I can search for ``Baupost Group founder'' to find the founder's name. Let me check that. \blue{</think>}
        \\\red{<search>} Baupost Group founder \red{</search>}
        \\\green{<result>}
        \\Page 1: ``Baupost Group'' ...and \$29.9 billion as of December 31, 2013. Baupost Group The Baupost Group is a long-only hedge fund founded in 1982 by Harvard Professor William Poorvu and partners Howard Stevenson, Jordan Baruch and Isaac Auerbach. Seth Klarman, who was asked by Poorvu to help run the fund, remains at its head today. Baupost Group's investment philosophy emphasizes risk management. The firm, one of the largest hedge funds in the world, is a value investing manager. According to Bloomberg L.P., Baupost is ranked 4th in net gains since inception.
        \\Page 2: ``Baupost Group'' Baupost Group The Baupost Group is a long-only hedge fund founded in 1982 by Harvard Professor William Poorvu... Seth Klarman, who was asked by Poorvu to help run the fund, remains at its head today.
        \\Page 3: ``Baupost Group'' ...Baupost CEO, Seth Klarman, explained in a speech to MIT students that investment research driven by emotion is risky and can lead to a bad investment. From its founding the firm's three private partnerships have generated an average annual return of 19\%.
        \\\green{</result>}
        \\\blue{<think>} So, from the search result, the Baupost Group was founded by William Poorvu and others. Seth Klarman was asked by Poorvu to help run the fund and remains at its head. The question is asking who the Baupost Group asked to run the hedge fund. The answer is Seth Klarman. Let me confirm: the search results consistently mention that Seth Klarman is the current head, and he was asked by Poorvu to help run the fund. So the answer should be Seth Klarman. \blue{</think>}
        \\\purple{<answer>} The final answer is \textbackslash boxed\{Seth Klarman\}. \purple{</answer>} \\
    \midrule
        \rowcolor{red!5}\textbf{No-tool path} (TC=0, EM=0.0) \\
        \blue{<think>} Okay, let's see. The question is about the Baupost Group asking which American investor, hedge fund manager, and philanthropist to run the hedge fund. Hmm. I remember that Baupost Group is a hedge fund, right? So they might have had a managing director or someone else appointed. Wait, I think the answer is someone like Warren Buffett. No, wait, Buffett is the CEO of Berkshire Hathaway. Maybe it's another person. Let me think. There's a person named Bill Ackman? No, he's more of an activist investor. Or maybe Roger L. Ivelt? No, that's not right. Wait, I think the correct answer is Dick Fuld. No, he was the CEO of Lehman Brothers. That's not it. Maybe someone else. The Baupost Group was founded by Robert F. Draper and Jerry Rhoads. But the question is about who they asked to run the fund. Perhaps the answer is Warren Buffett after all. \blue{</think>}
        \\\purple{<answer>} The final answer is \textbackslash boxed\{Warren Buffett\}. \purple{</answer>} \\
    \bottomrule
    \end{tabular}
\end{table*}

\begin{table*}[!t]
    \centering
    \caption{Case \#3: \textbf{Hallucination} ($WT$=\ding{55}, $NT$=\checkmark). The retriever returns irrelevant results that mislead the model into an incorrect answer. The no-tool trajectory is selected as the target signal.}
    \fontsize{8.5pt}{10.5pt}\selectfont
    \begin{tabular}{@{}p{\textwidth}@{}}
    \toprule
        \textbf{Question:} Who died first, Maurice Pialat or Arline Burks Gant? \quad \textbf{Ground Truth:} Maurice Pialat \\
    \midrule
        \rowcolor{red!5}\textbf{With-tool path} (TC=4, EM=0.0) \\
        \blue{<think>} Okay, so the question is asking who died first between Maurice Pialat and Arline Burks Gant. I need to find the dates of their deaths and compare them. First, I should look up when Maurice Pialat died. Let me search for that. \blue{</think>}
        \\\red{<search>} Maurice Pialat death date \red{</search>}
        \\\green{<result>}
        \\Page 1: ``Maurice Pialat'' Maurice Pialat (; 31 August 1925 -- 11 January 2003) was a French film director, screenwriter and actor noted for the rigorous and unsentimental style of his films. His work is often described as being ``realist''. Pialat was born in Cunlhat, Puy-de-D\^ome, France.
        \\Page 2: ``Maurice Pialat'' ...He was always willing to bend his narratives around experience... Maurice Pialat (; 31 August 1925 -- 11 January 2003) was a French film director...
        \\Page 3: ``Maurice Ronet'' ...He died in a Paris hospital, of cancer... Maurice Ronet (13 April 1927 -- 14 March 1983) was a French film actor, director, and writer.
        \\\green{</result>}
        \\\blue{<think>} So Maurice Pialat died on 11 January 2003. Now, I need to find when Arline Burks Gant died. Let me look that up. \blue{</think>}
        \\\red{<search>} Arline Burks Gant death date \red{</search>}
        \\\green{<result>}
        \\Page 1: ``Barbara Stoddard Burks'' ...years later. Burks married Herman Ramsperger, a National Research Fellow in chemistry at Stanford, in 1927; they remained married until his death in 1932. In 1943, she became engaged to longtime friend Robert Cook, but she died before they could be married. Burks died on May 25, 1943, when she ``either fell or jumped to her death from the George Washington Bridge'' in New York City.
        \\Page 2: ``Richard Fenner Burges'' ...he was married to Ethel Petrie Shelton... He died in El Paso on January 13, 1945.
        \\Page 3: ``Abbie G. Rogers'' ...died unexpectedly on May 21, 1894, age 53 in New York City. She had been undergoing an operation to remove a tumor.
        \\\green{</result>}
        \\\blue{<think>} Hmm, the first result mentions Barbara Stoddard Burks, but that's a different person from Arline Burks Gant. The second result is about Richard Fenner Burges. The third is about Abbie G. Rogers. None of these are about Arline Burks Gant. Wait, maybe Arline Burks Gant is the same as Barbara Stoddard Burks? The name ``Burks'' appears in both. Alternatively, maybe there's a mix-up. Assuming that Arline Burks Gant is related to the ``Burks'' who died in 1943, then she died before Maurice Pialat (2003). Therefore, Arline Burks Gant died first. \blue{</think>}
        \\\purple{<answer>} The final answer is \textbackslash boxed\{Arline Burks Gant\}. \purple{</answer>} \\
    \midrule
        \rowcolor{green!5}\textbf{No-tool path} (TC=0, EM=1.0) $\leftarrow$ \textit{Selected as target} \\
        \blue{<think>} The question asks who died first between Maurice Pialat and Arline Burks Gant. I recall that Maurice Pialat was a well-known French film director who passed away on January 11, 2003. As for Arline Burks Gant, she is less well-known, but I believe she lived until 2011. Therefore, Maurice Pialat died first in 2003, before Arline Burks Gant. \blue{</think>}
        \\\purple{<answer>} The final answer is \textbackslash boxed\{Maurice Pialat\}. \purple{</answer>} \\
    \bottomrule
    \end{tabular}
\end{table*}

\end{document}